\documentclass[10pt,twocolumn,letterpaper]{article}

\usepackage{cvpr}
\usepackage{times}
\usepackage{epsfig}
\usepackage{graphicx}
\usepackage{amsmath}
\usepackage{amssymb}

\usepackage[breaklinks=true,bookmarks=false]{hyperref}

\cvprfinalcopy 

\def\cvprPaperID{****} 

\begin{document}
	
	\title{\textbf{Learning without Prejudice: Avoiding Bias in Webly-Supervised Action Recognition}}

\author{Christian Rupprecht$^{1,2}$
	\and
	Ansh Kapil$^{1}$
	\and
	Nan Liu$^{1}$
	\and
	Lamberto Ballan$^{3}$
	\and
	Federico Tombari$^{1}$
	\and
	\\$^{1}$Technische Universit\"at M\"unchen
	\and
	\\$^{2}$Johns Hopkins University
	\and
    \\$^{3}$University of Padova}


\maketitle

\begin{abstract}
Webly-supervised learning has recently emerged as an alternative paradigm to traditional supervised learning based on large-scale datasets with manual annotations.
The key idea is that models such as CNNs can be learned from the noisy visual data available on the web.
In this work we aim to exploit web data for video understanding tasks such as action recognition and detection. One of the main problems in webly-supervised learning is cleaning the noisy labeled data from the web. The state-of-the-art paradigm relies on training a first classifier on noisy data that is then used to clean the remaining dataset. Our key insight is that this procedure biases the second classifier towards samples that the first one understands.
Here we train two independent CNNs, a RGB network on web images and video frames and a second network using temporal information from optical flow. 
We show that training the networks independently is vastly superior to selecting the frames for the flow classifier by using our RGB network.
Moreover, we show benefits in enriching the training set with different data sources from heterogeneous public web databases. 
We demonstrate that our framework outperforms all other webly-supervised methods on two public benchmarks, UCF-101 and Thumos'14.
\end{abstract}




\section{Introduction}
In recent years, deep learning has fueled a significant progress in several computer vision tasks. One of the main reasons behind such achievements is the development of large-scale datasets with annotations \cite{imagenet_cvpr09,imagenet,lin2014coco,krishnavisualgenome}, which enable training of deep neural networks with millions of parameters without over-fitting.
This has led deep learning based models to approach human level performance on various visual data classification and recognition tasks \cite{imagenet}.
However, data annotation has intrinsic limitations -- both in terms of time and cost. This is even more critical for video data, since annotating action labels and defining temporal bounds for thousands of videos is particularly tedious and time consuming, which makes it a non-scalable solution. Moreover, this manual annotation process often introduces a bias towards very specific tasks and domains \cite{Ponce2006,unbiased}.  

\begin{figure}[!t]
	\centering
	\includegraphics[width=0.9\linewidth]{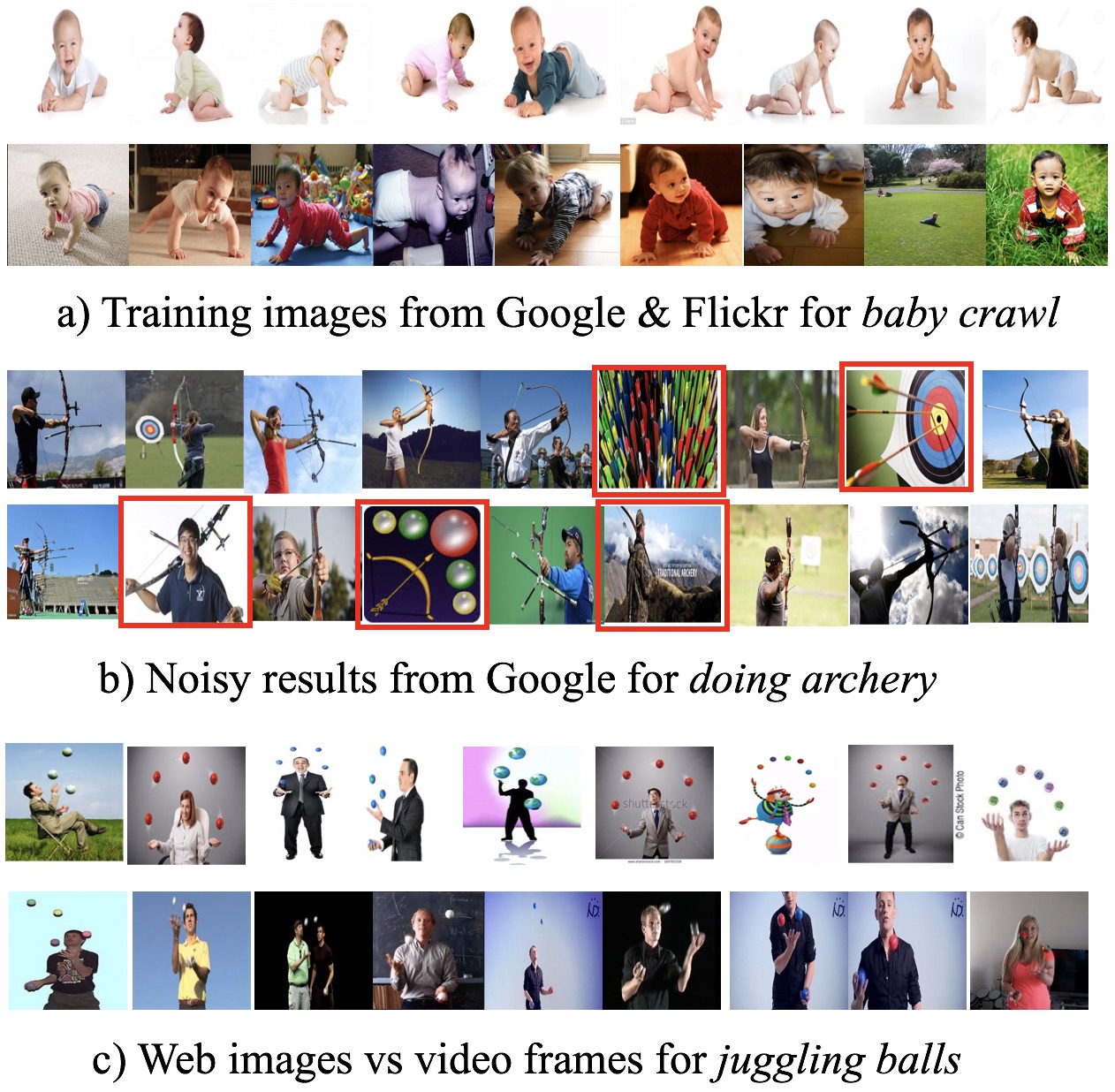}
	\caption{Web images and videos collected from heterogeneous web sources are characterized by different appearance and noise. In this work we present a fully \emph{webly-supervised} approach for recognizing and localizing a large number of action categories in trimmed and un-trimmed videos.}
	\label{fig:teaser}
\end{figure}

To overcome these limitations, the \emph{webly-supervised} paradigm has emerged as an appealing alternative which aims at learning features and training models by solely relying on noisy web data \cite{optimol,chen2013neil,divvala2014levan,zhou2015,joulin2016web}.
The clear advantage is that the amount of data obtainable from the Internet is huge and continuously growing, and vastly exceeds what is achievable through manual annotation.
Notably, webly-supervised approaches already perform competitively to the state-of-the-art for certain visual recognition tasks, such as image classification and object detection. For example, in their recent paper Chen and Gupta \cite{chen2015webcnn} show that a CNN trained only with web images gives comparable performance to the ImageNet \cite{imagenet_cvpr09} pre-trained network architecture for object detection, without using a single manually annotated label.
Similarly, Joulin \etal\cite{joulin2016web} show that it is possible to train CNNs on massive weakly-labeled image collections -- such as the 100 million Flickr images dataset \cite{YFCC100M} -- for learning good visual representations.

Inspired by this line of work, we propose a CNN-based webly-supervised method for the task of human action recognition from real world videos, where the visual data is entirely collected from the web.
This task has only recently started to be addressed in literature \cite{sun2015,gan2016webly,gan2016you} and poses several challenges.
First, the labels obtained through web search queries are often noisy and do not represent the content -- in terms of actions -- of the retrieved video. Hence, a proper filtering has to be adopted to remove outliers from the samples that will be used as training set.
Second, a key challenge is represented by \emph{untrimmed} videos. Conversely to their \emph{trimmed} counterpart, which contain one single action for their whole length, these videos also include several background frames, i.e. without any human activity. 
Often, such videos are long and might contain multiple actions, belonging to either the same or even different classes.
To overcome such challenges, a common strategy is to train an initial set of classifiers from a particular source (e.g. images) and use the classifier scores to filter out noise and represent the other source (e.g. videos) \cite{ikizlercinbisICCV2009,chen2014dvmm,singh2015web,gan2016webly}.
A good example of this approach is the recent work of Gan \etal \cite{gan2016webly} in which video concepts are discovered using two separate classifiers. The first one is trained on noisy data obtained from a specific web source (i.e. YouTube videos), then it is  used to clean the samples used to train the second classifier (trained on Google images).

Our approach stems from the key insight that this procedure biases the second classifier towards samples that the first one understands well, since it subjectively limits the variability of the training set. The proposed idea is thus to train two independent classifiers -- in particular, two CNNs -- one specialized on web images and video frames, the other one encompassing temporal information obtained from optical flow computed on subsequent video frames.
We demonstrate that the use of two independent networks can yield significant benefits in terms of classification accuracy, rather than having the image-based classifier selecting the frames for the optical flow-based classifier. 
In addition, we also improve the generalization ability of the networks, since most action images retrieved with Google Image Search are usually background free with the human in the center.
The idea is thus to enrich the training set with different heterogeneous web data sources. 
To this end, we include images from Flickr, as well as frames collected from videos available on Youtube, the latter being particularly useful also to reduce the \emph{semantic gap} between the actions being sought and the images downloaded from the web.

The proposed approach is tested on two publicly available datasets, namely UCF-101 \cite{UCF101} and Thumos'14 \cite{THUMOS14}, demonstrating state-of-the-art results with respect to other webly-supervised approaches, as well as a performance comparable to several recent approaches trained with clean, manually annotated datasets.

\section{Related work}
There is a long history of research in video event detection and action recognition. Thorough related surveys are given in \cite{poppe2010survey,weinland2010survey,ballan2011survey}.

\paragraph{Video Recognition and Action Understanding}
Early work focused on defining hand-crafted features for video analysis such as 3D Histogram of Oriented Gradients (with the third dimension being time) and spatio-temporal features based on optical-flow \cite{dollar2005behavior,klaser2008spatio,laptev2008learning}.
State-of-the-art hand-crafted features, i.e. the dense trajectories of Wang \etal\cite{wang2013dense}, achieved excellent results on multiple action recognition benchmarks \cite{peng2014action,wang2013action}. 
However, handcrafting features requires a rich domain knowledge. In real-world videos, the construction of such features may vary consistently among different domains, and hence application specific features might be required \cite{lan2015beyond,de2016sympathy}.

Following the success of deep learning architectures in image classification \cite{krizhevsky2012imagenet,imagenet}, there have been attempts to extend this paradigm to videos as well. To this end, \cite{ji20133d} proposes a technique to extend CNNs for video analysis, where features are learned from both, spatial and temporal dimensions by using 3D convolutions. 
Karpathy \etal\cite{karpathy2014large} and Tran \etal\cite{tran2015learning} learn spatio-temporal filters and employed different pooling schemes across the temporal axis to address time.
Simonyan and Zisserman \cite{simonyan2014twos} propose a two-stream convolutional network for action recognition in which they learn separate CNNs for different input sources. First, they train a model on RGB video frames for spatial features, and then they use optical flows between video frames to incorporate motion information. 
Temporal segment network have been recently proposed to overcome the limitation of these two-stream CNNs in modeling long-range temporal structure \cite{wang2016tsn}.
However, the common setup is to rely on a large dataset annotated by human experts in a fully supervised setting.

\begin{figure*}[!t]
\centering
\includegraphics[width=0.95\linewidth]{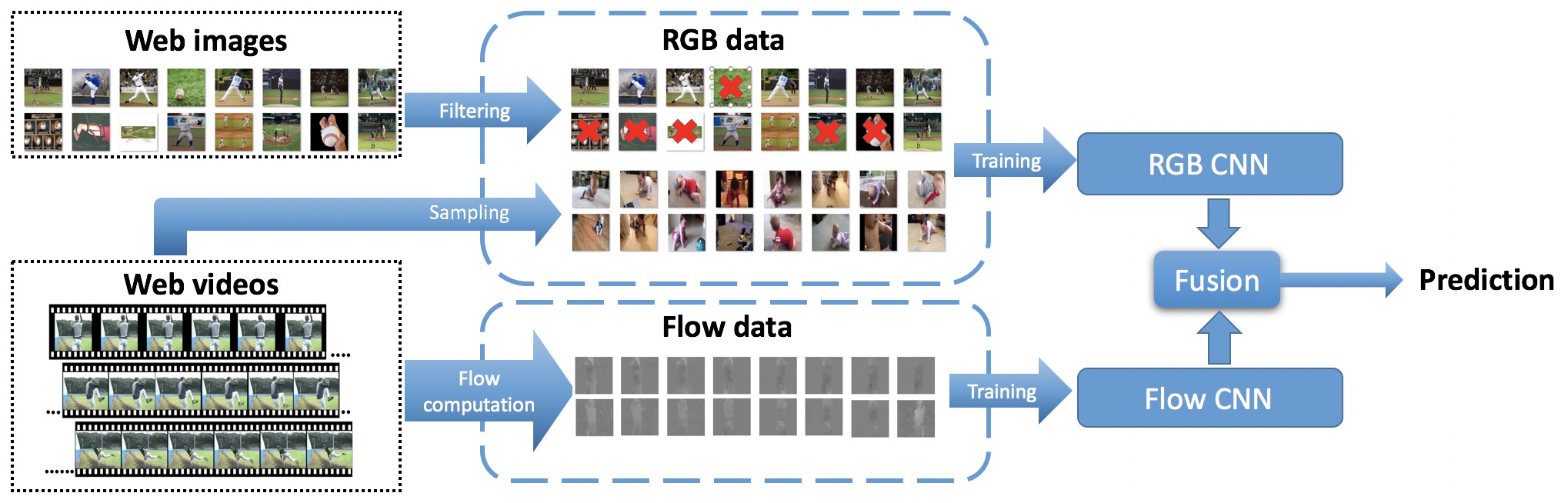}
\caption{Overview of the proposed model.}
\label{fig:model_overview}
\end{figure*}

\paragraph{Learning from Web supervision}
Previous work for learning visual knowledge from the web typically focuses on images and addresses tasks such as image classification and object detection \cite{optimol,chen2013neil,divvala2014levan,krause2016web}.
Recently, some works have also proposed to learn CNNs \cite{joulin2016web, chen2015webcnn} and visual concepts \cite{zhou2015,li2013midconcept,sun2015automatic} from noisy web data.
This is also closely related to the vast literature on image and video tagging \cite{li2016socializing,toderici2011youtube,cviu2015videos}.
Inspired by this line of work, we focus on the specific problem of action recognition in large-scale video data.

There have been few attempts to tackle video understanding tasks using a webly-supervised approach.
In their seminal work, Ikizler \etal~\cite{ikizlercinbisICCV2009} present a model for action recognition in videos using web data. After collecting images from the web, a person detector is used to filter out noisy images, then a linear classifier is learned on the resulting features and applied to the test videos.
Similarly, in \cite{chen2014dvmm,singh2015web} a set of concept detectors is discovered from the web using Google or Flickr images. After this preliminary learning step, these concept models are applied to the videos to generate video-level representations, which can be used in a fully supervised setting for event detection.
More closely related to our work is the recent ``Lead-Exceed'' Neural Network presented in~\cite{gan2016you}, where a CNN network trained on web videos is refined using images collected from the web in a curriculum learning manner. 
Similarly,~\cite{sun2015} uses a domain transfer scheme between images and videos to filter out the noise in the other domain, and obtain so-called Localized Actions Frames (LAFs). An LSTM over LAFs is then applied for fine-grained action recognition and localization. The localization problem is also tackled in~\cite{sultani2016cvpr}, where they present a weakly-supervised model to perform spatio-temporal localization in videos.
However, in these approaches the proposed procedure biases the recognition process towards samples that are recognized in the first domain or modality.

\section{Proposed Approach}
In this work we exploit web images and videos for action recognition and localization in a fully webly-supervised fashion. Figure \ref{fig:model_overview} shows our model structure.
An important observation is that different data sources provide varied types of data and labels. For example, images retrieved from Google image search are usually clean, with the object of interest centered in the image and a monochrome background. In contrast, images from web sources like Flickr are more natural, in the sense that the object is usually depicted \textit{in the wild}. This discrepancy between different sources produces an effect that we dub as the \emph{source bias}.
Another biasing factor arises when using a supervised classifier to filter outliers to determine a training set for a successive classifier (like in \cite{ikizlercinbisICCV2009,chen2014dvmm,singh2015web,gan2016webly}).
In this case, the bias is introduced by the fact that the ``inliers'' determined by the first classifier will be only samples that were well understood by the first classifier. We refer to this problem as \emph{filter bias}.

We propose a methodology that avoids both biasing factors, respectively by employing the following strategies:
\begin{itemize}
	\item \textbf{Source Bias}: mix data from different web sources to reduce the source bias introduced by prevailing image structures typical for a single data provider.
	\item \textbf{Filter Bias}: avoid using a supervised procedure to filter outliers so not to introduce bias through a specific training set.
\end{itemize}

Our action recognition pipeline consists of two main components: \emph{i}) data collection and filtering (Section \ref{data_collection} and \ref{data_filtering}). \emph{ii}) training a two-stream CNN architecture, one stream based on RGB data and the other based on optical flow, similar to \cite{simonyan2014twos} (Sections  \ref{rgb_net} and \ref{flow_net}). 

We summarize the proposed pipeline in four steps: \emph{i}) download a set of web images $\mathcal{I}$ and a set of web videos $\mathcal{V}$ for each action label; \emph{ii}) filter $\mathcal{I}$ and $\mathcal{V}$ to obtain $\hat{\mathcal{I}}$ and $\hat{\mathcal{V}}$; \emph{iii}) train a network on RGB frames from the filtered sets $\hat{\mathcal{I}}$ and $\hat{\mathcal{V}}$; \emph{iv}) train a network on stacks of optical flow maps from the set $\hat{\mathcal{V}}$. We detail the individual steps of our pipeline in the following subsections.

\subsection{Data Collection}\label{data_collection}
Google search often returns images in which the human actor is located in the image center in front of a uniform background. Training with this data can lead the model to under-perform in cluttered scenes. To resolve this, we add real world images chosen from Flickr, which are more diverse in terms of actor position and background scene. Relevantly, Flickr has been used for many benchmark datasets like Pascal VOC \cite{Everingham10}. 
Additionally, web searches for some classes return results which do not represent an action in the wild. There is a so called \textit{semantic gap} \cite{smeulders2000content,li2016socializing} in how the action appears in real world videos and how they are depicted in web images. This semantic gap can be reduced by using RGB frames from web videos which we collect from YouTube. As in the case of image data, YouTube is queried with the action names. 
Finally, since web videos are often rather long and the action can appear in any part of the sequence, we add action videos as animated GIF files by querying Google search by activating the ``animation result only'' filter, as well as Giphy (\url{giphy.com}), an online GIF database to download GIF files for the action. Due to poor compression inherent in the file format, GIFs are usually very short, this in turn increasing the probability that the majority of a sequence will represent the desired action.

Similar to Gan \etal\cite{gan2016you}, we perform small changes to the category labels for querying the web. For example for the label \texttt{nun chucks}, the search term is modified to \texttt{doing nun chucks} to avoid retrieving pictures and videos of the object instead of the corresponding action.

\subsection{Filtering}\label{data_filtering}
\begin{figure}[t]
	\centering
	\includegraphics[width=\linewidth]{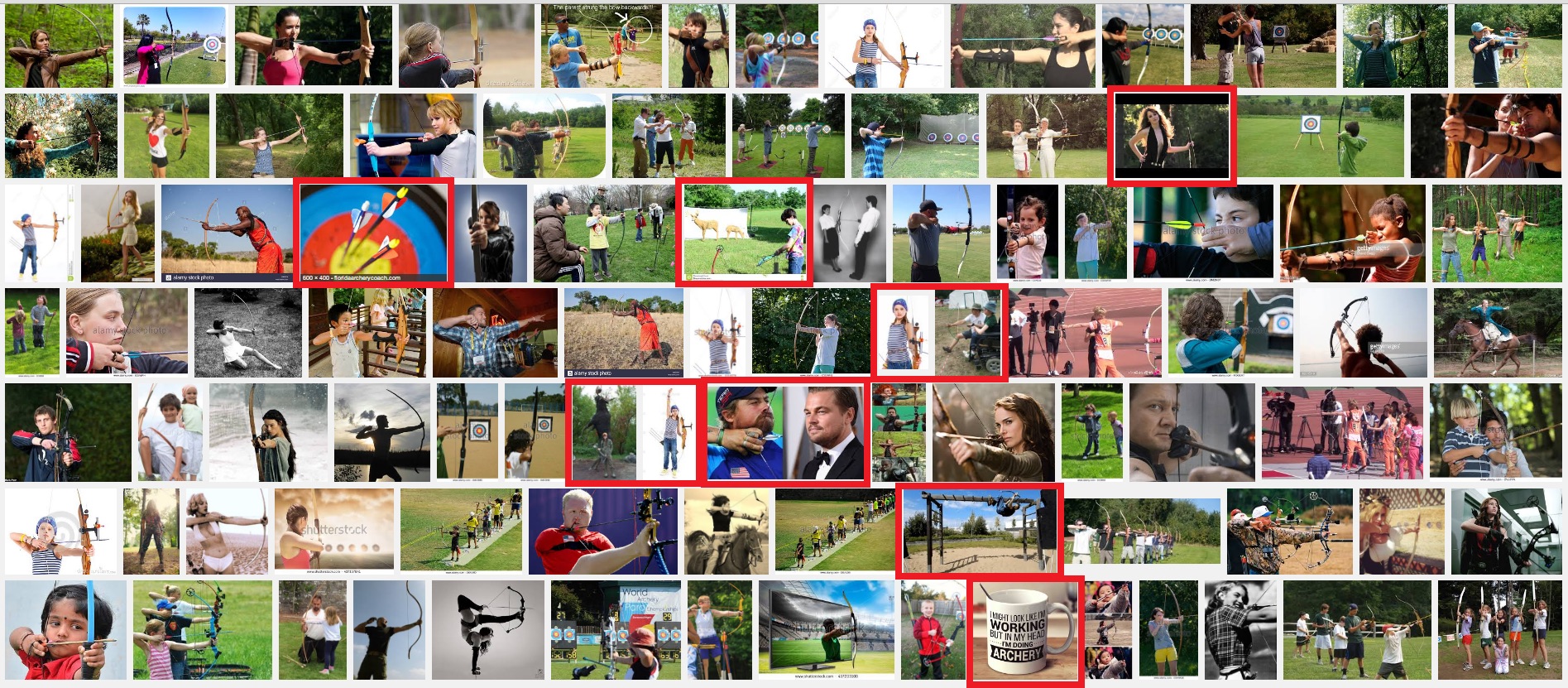}
	\caption{\textbf{Query:} \texttt{doing archery} (from images.google.com). We highlight all the frames that we do not expect to encounter in a video depicting a human \texttt{doing archery} with a red border.}
	\label{fig:noisysearch_archery}
\end{figure}

The images collected from the web contain outliers that do not belong to the query, such as different action classes, animations, or just text. To minimize the influence of these samples it is necessary to filter the retrieved images before training a classifier. 
To portray an example, Figure \ref{fig:noisysearch_archery} shows the image search results given the query \texttt{doing archery} from Google image search. Clearly, the images highlighted in the red box are outliers since they do not depict any human action, and will confuse the model being learned if included in the training set. 

As previously mentioned, we aim at filtering the image set in a way that prevents the aforementioned filter bias. Filter bias can formally described as a set $\mathcal{C} \subset \mathcal{X}$ of correctly labeled samples being corrupted by outliers $\mathcal{O} \subset \mathcal{X}$. We are given the superset $\mathcal{S} = \mathcal{C} \cup \mathcal{O}$ and need to find a filter function $f: \mathcal{X} \to \{-1,1\}$ that classifies elements of $\mathcal{C}$ as positive and detects outliers from $\mathcal{O}$ as negative such that the selected set $\hat{\mathcal{C}} = \{x | f(x) = 1, x \in \mathcal{S}\}$ equals the clean samples $\hat{\mathcal{C}} = \mathcal{C}$. Hence, filter bias occurs when $f$ selects a strict subset of $\hat{\mathcal{C}} \subset \mathcal{S}$ which changes the distribution of samples and thus biases any classifier trained on it.

A common strategy deployed by state-of-the-art webly-supervised approaches for action recognition ~\cite{ikizlercinbisICCV2009,sun2015,gan2016you} is to train a classifier that is then used to filter out noise on the remaining set. We will hereinafter refer to this paradigm as \emph{supervised filtering}. 
As we will show in the experimental section, such supervised filtering approaches easily lead to filter bias, as they tend to filter out difficult but correct samples, thus biasing the resulting set $\hat{\mathcal{C}}$ to contain many simple examples and only few difficult ones. 

\begin{figure}[t]
	\centering
	\includegraphics[width=\linewidth]{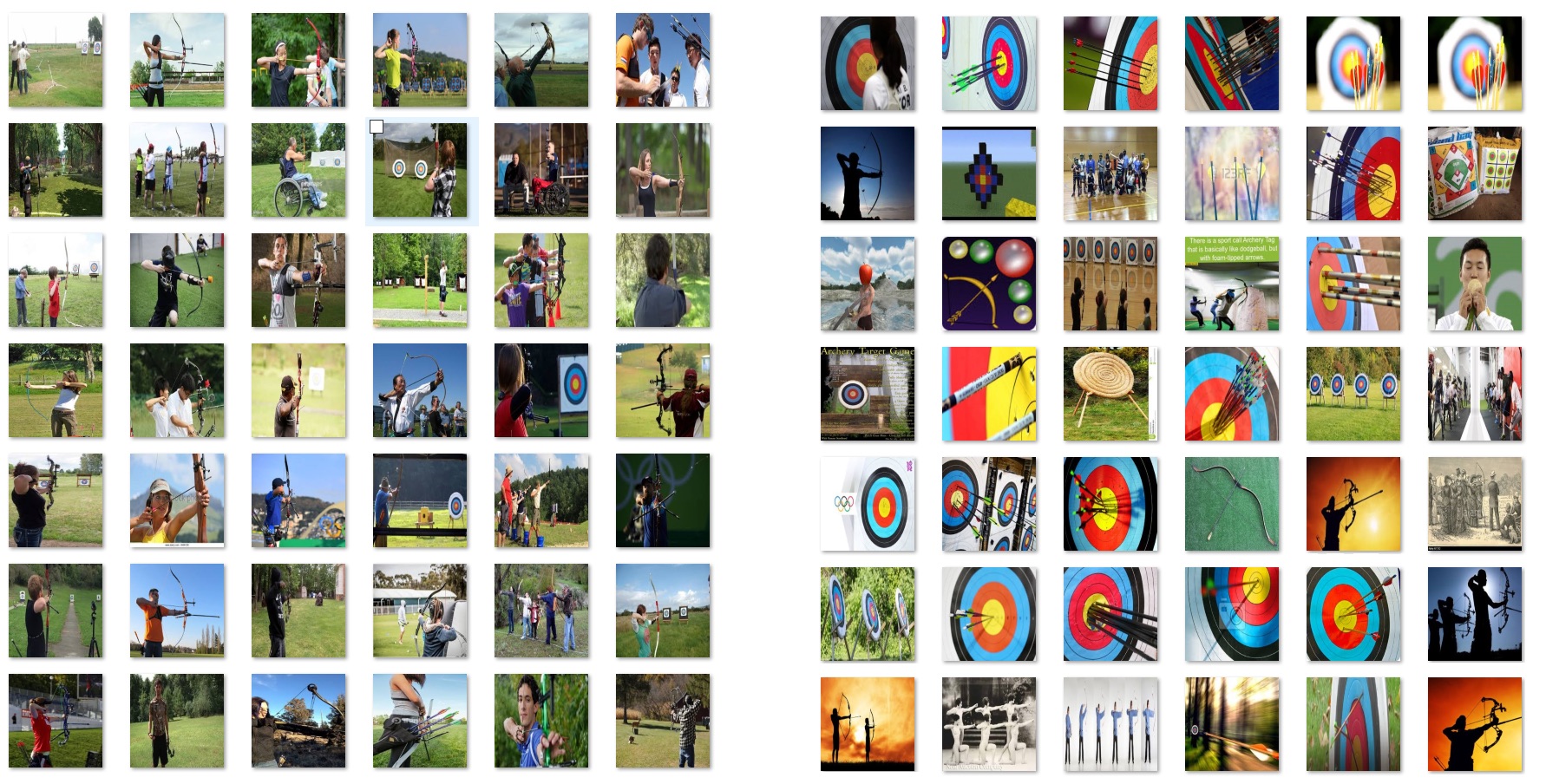}
	\caption{Images of action \texttt{archery} after applying random walk filtering. \textbf{Left:} image subset obtained by removing the noisy images. \textbf{Right:} images filtered out by random walk. }
	\label{fig:randomwalk_archery}
\end{figure}

Hence, we suggest to use an approach for filtering web data outliers that does not rely on a training set. We refer to this filtering paradigm as \emph{independent filtering}. Specifically, we rely on a filtering algorithm based on random walk similarly to~\cite{moonesignhe2006outlier,sultani2016cvpr}.
To this end, we define a fully connected graph $\boldmath Z(N,E)$, where $N$ is the set of all $n$ images and $E$ represents the set of edges between them. We map an image $N_i$ to a feature vector using $\phi(N_i)$. The Euclidean distance in feature space of a pair of nodes $\phi(N_i)$ and $\phi(N_j)$ is a measure of similarity. A small distance implies similar images.
The transition probability between any two nodes $N_i$ and $N_j$ is given by:
\begin{equation} 
p_{i,j} = \frac{e^{- \gamma \|\phi(N_i) - \phi(N_j)\|_2}}{\sum_{m=1}^n e^{-\gamma \|\phi(N_i) - \phi(N_m)\|_2}} 
\end{equation}
We will compute the relevance $r_k(N_j)$ of image $N_j$ iteratively over iterations $k$. Let $v_j = \frac{1}{n}$ be the initial probabilistic score. The update rule can then be written as:
\begin{equation}
r_k(j) = \beta \sum_i^n r_{k-1}(i)p_{i,j} + (1-\beta)v_j
\end{equation}
where $\beta$ controls the contribution of both terms to the final score. In all our experiments we set $\beta = 0.99$ and $\gamma = 0.01$.
The filtered image set for the \texttt{archery} action are shown in Figure \ref{fig:randomwalk_archery}. Most images which do not contain any human action such as target boards, arrow set, clip-art images are sorted out.

\subsection{RGB Network}\label{rgb_net}
As for the RGB network, four different models were trained on color frames to evaluate the contribution provided by different web sources: 
\begin{itemize}
	\item \textbf{Web images:} trained with only web images collected from Google and Flickr. After filtering with random walk, the best 450 images are selected for training.
	\item \textbf{Web videos:} per class, we sample 50 videos from YouTube to obtain 1500 frames. After random walk, the 500 highest ranked frames were chosen for the training set.
	\item \textbf{Web images + videos:} the training set includes 400 web images, 500 frames from videos and 100 additionally frames from animated GIFs, for a total of 1000 training samples per class. We train two variants: one's training set is filtered by random walk, the other's by the network trained only on web images.	
\end{itemize}
For all networks, the images are randomly separated into a training ($80\%$) and a validation set ($20\%$). The network, a 50-layer ResNet \cite{DBLP:journals/corr/HeZRS15}, is trained using Caffe \cite{jia2014caffe} with pre-trained weights from ImageNet \cite{imagenet_cvpr09}. Stochastic Gradient Descent (SGD) with a batch-size of 10 and a learning rate of $10^{-5}$ (decreased 10 times after every 100k iterations) is used. After 200k iterations the optimization converges.

\subsection{Optical Flow Network}\label{flow_net}
Motion is an important cue for the identification of actions. When the models are trained only with still RGB images, the lack of temporal information affects performance. In this section, the focus is on adding temporal motion information to the model. We train a CNN using optical flow images as input to predict the action class. Optical flow between pairs of consecutive video frames represents a short motion. To capture longer temporal dependencies, the optical flow images are stacked for a sequence of frames. Such inputs implicitly describe the motion in a sequence, which makes recognition easier.

Similar to RGB images, the input for the network in case of optical flow is also in the form of a volume. The flow images are split into horizontal ($x$) and vertical ($y$) displacements. These displacements are stacked one after another as multiple channels, with the goal of modeling longer temporal dependencies. For frames of size $w \times h$, the input volume for the network will be $w \times h \times 2D$. $D$ represents the number of stacked optical flow frames.

For the flow classification model, the feature map of the first convolutional layer has size $64\times3\times7\times7$ since it was trained on 3-channel RGB images. The mean of the weights across the 3 channels can be replicated $2D$ times to match the new input dimensions. The resultant first layer will thus have a size of $64\times2D\times7\times7$. The weights of all remaining layers are kept the same for initialization to train a Resnet-50 model~\cite{DBLP:journals/corr/HeZRS15}.
Again, the solver is SGD with a batch size of 10. The training is performed over 60 epochs and the initial learning rate is set to $10^{-3}$ which is reduced by a factor of 10 after every 20 epochs. Brox's method ~\cite{brox2004high} is used to compute optical flow between two consecutive frames.

Each frame sequence is run separately through the RGB and the flow network to obtain video level probabilities. The probabilities from the two networks are then combined by two different schemes, namely \emph{fusion-by-averaging} and \emph{fusion-by-product}. In the first case, we compute the element-wise average of the two probability vectors from the RGB and the flow CNN. In the second case, instead, we compute the element-wise product of the two probability vectors from the two networks. The class with maximal probability forms the final prediction.

\subsection{Action Understanding Tasks}
A major benefit of the proposed pipeline is that the trained networks can be used for three different action understanding tasks. We show how the networks, trained on the same set of action labels, can be used for trimmed action classification, untrimmed action classification and action localization. To the best of our knowledge, this is the first work that employs the same fully webly-supervised method for all three tasks. 

Action classification in trimmed videos is arguably the easiest of the three tasks, since at test time the algorithm is given a short video that fully contains a single action to be classified. Since the action covers the whole clip, there is no presence of background actions which might confuse the network. 
To compute a label for each video, the probability vectors corresponding to each frame, obtained by forward pass through the trained CNNs, are averaged along the temporal axis to get the final score for the video. Differently, action classification in untrimmed videos includes the additional challenge that the video is not cut around the action. Videos are typically longer and the action to be recognized spans only a short portion of the whole clip. In addition, the action to be recognized could be present in one or multiple instances. Although the system is trained only on actions, it needs to be robust to long background sequences in between actions. Here, we also average all frame-wise probability scores temporally to yield the score for each video. 

Finally, untrimmed action localization is the task where, in addition to the action label, temporal action boundaries defining the start and end moments of the action in the video have to be estimated. 
This problem is particularly challenging since the background frames often bear some resemblance to the action. For instance, most videos with activity labels such as "diving", "breast stroke", "front crawl" are all similar to each other, as they contain a swimming pool. We employ two different techniques to perform the temporal localization of actions. As for the first one, referred to as \emph{frame-by-frame} localization scheme, the global level prediction for the whole video is first obtained from the network. Then, the action is localized on a frame-by-frame basis. Specifically, all video frames conforming to the global prediction above a certain probability threshold are grouped together along the temporal axis. Such sequences that are longer than 0.1 seconds count as positive localizations.
The second localization scheme, denoted as \emph{sliding window}, uses a small sub-window of the full video and tries to localize the action therein. Differently to \emph{frame-by-frame} localization, no global level predictions or thresholds are taken into account. If an action is predicted in the sub-window by averaging its single predictions, the whole window's temporal bound is reported.

\section{Results}
We benchmark our framework on two publicly available large-scale datasets, UCF-101~\cite{UCF101} and Thumos'14~\cite{THUMOS14}, and compare to the state of the art.

\subsection{Datasets and Metrics}
\textbf{UCF-101}~\cite{UCF101} is a large-scale dataset consisting of 13,320 trimmed videos from 101 action classes. We test our webly-supervised model on the provided three test splits consisting of 3783, 3734 and 3696 video respectively. The metric used to measure performance is classification accuracy (Acc.) averaged across the three test splits.

\textbf{Thumos'14}~\cite{THUMOS14} is a large scale dataset consisting of only untrimmed videos from the same 101 action classes as of UCF-101. An untrimmed video may contain one or multiple instances of same or different actions within its temporal bounds. This dataset presents two tasks: (1) action recognition in video, as well as (2) localizing the action temporally among the videos.
The test set consists of 1574 videos from 101 action classes, but the localization task is only applicable for 20 selected classes.
The metric used for evaluation as per official Thumos'14 protocol is mean Average Precision (mAP).

\subsection{Analysis of Filtering Method}
\begin{figure}[t]
	\centering
	\includegraphics[width=0.95\linewidth]{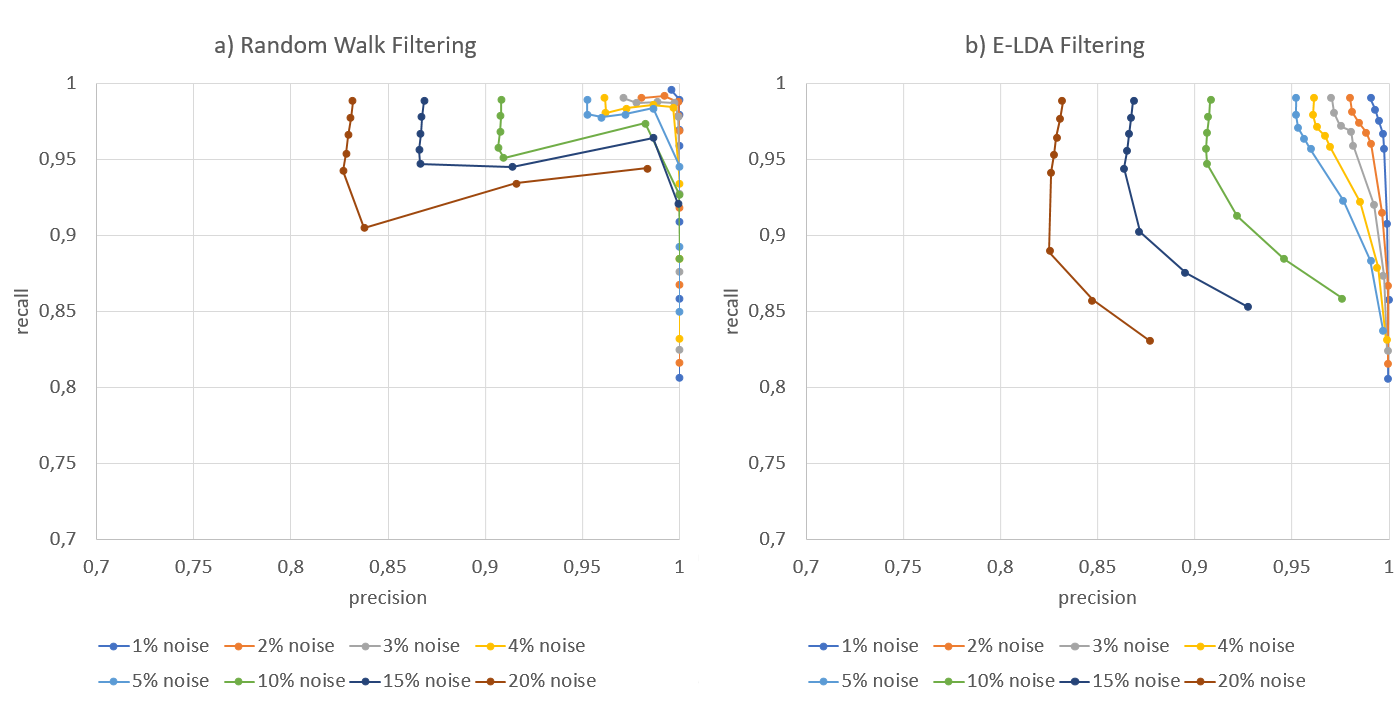}
	\caption{Analysis of filtering methods on the UCF-101 dataset. a) shows precision-recall curves filtering results for the Random Walk based method while b) depicts the results for E-LDA. In all noise levels RW filtering performs better than E-LDA.}
	\label{fig:filtering}
\end{figure}

In this section we compare the random walks filtering technique to the E-LDA \cite{hariharan2012discriminative} based filtering of \cite{chen2015webcnn}. To be able to use ground truth for this task, we use images from the PASCAL VOC 2012 dataset, which in turn was created from web (flickr) images. For all our experiments we take all images of one of the 20 classes and subsequently add more and more 'noisy samples' in the form of images from the other classes. We then run E-LDA and RW filtering on the corrupted dataset and can precisely measure how well the outliers are removed by the filtering procedure. The experiment is repeated and averaged over 19 classes\footnote{Class \textit{person} was excluded due to the much larger amount of images which would take more than one week for the noise analysis for E-LDA.}.
To the original image we add $1\%$, $2\%$, $3\%$, $4\%$, $5\%$, $10\%$, $15\%$ and $20\%$ noise samples and then filter with the same thresholds in all noise levels. As expected the best performance in achieved when the noise level matches the filtering amount. When more is filtered than actual noise, naturally the recall decreases while precision stays maximal.

\begin{table}
	\centering
	\begin{tabular}{  l | r  }\hline
		\textbf{Network trained on} & \textbf{Accuracy \%} \\ \hline
		web video frames & 52.40 \\	
		web images & 62.44 \\
		video frames + images & \textbf{65.60} \\ \hline
	\end{tabular}
	\caption{Source Bias experiment: performance on the UCF-101 dataset of different networks trained on RGB data only.}
	\label{table:RGBNets}
\end{table}

The results can be found in Figure \ref{fig:filtering}. The Figure clearly shows that RW filtering achieves a better performance than E-LDA at all noise levels. Furthermore, it is noteworthy that RW-filtering with up to 15\% noise always reaches 100\% precision with over 90\% recall. This ensures that no false positives are left in the data and the network can be trained with only clean images. It is also interesting to note, that, especially for higher noise levels, the recall first drops and then the precision increases. This indicates that there are some outliers that are difficult to detect for both filtering techniques and are only removed by RW with a higher threshold. E-LDA is not able to identify the outliers especially with noise greater than 5\%.

\subsection{Evaluation of Bias Removal}
We will analyze the effects of the two different types of biases on our model. Firstly, we analyze the effects of source bias on UCF-101.
Table \ref{table:RGBNets} shows the benchmark results of training three networks. All training sets are filtered by random walk. When using web video frames only, we achieve the lowest performance since web videos are untrimmed and the desired action frames are often surrounded by large number of background frames leading to difficulties in removing outliers. On the other hand, using web images alone, we gain 10\% accuracy since there are usually less outliers in the search results. Finally, by combining the data of the two sources we achieve the best performance. This result clearly shows that combining different data sources improves the generalization ability of the trained classifier, as the source bias decreases.
 
\begin{table}
	\centering
	\begin{tabular}{  l| l | r }\hline
		\textbf{Input} & \textbf{Filter} &\textbf{Accuracy \%} \\ \hline
		Flow only & RGB net & 15.46 \\ 
		Flow only & Random Walk & \textbf{50.67} \\ \hline
	\end{tabular}
	\caption{Filter Bias experiment: optical flow classification performance by filtering the training set with random walk and the RGB frame classifier on UCF-101 (split 1).}
	\label{table:opticalFlowResults}
\end{table}

In a second experiment, we investigate the effects of filter bias. 
In Table \ref{table:opticalFlowResults} we compare the recognition accuracy for optical flow maps by comparing two filtering methods. We select RGB frames from the videos, compute its optical flow map and add it to the training set. We show two methods for RGB frame selection. One is by using the independent Random Walk filtering, the other uses RGB network from the previous experiments to select frames with high confidence. In the results one can clearly observe the degraded performance induced by filtering with the classifier. This is caused by two factors: the RGB network works well for video frames of actions that can be classified easily by a single frame. This is not necessarily true for optical flow maps. In fact, the whole idea behind using flow additionally to the images was to create two complementary classifiers. Pre-selecting the training data with the RGB network defeats this purpose. Secondly, for some classes the RGB network is almost never confident enough to generate sufficient training data for the flow network.

\begin{table}
	\parbox{\linewidth}{
		\centering
		\begin{tabular}{l|l|r}\hline
			\textbf{Training Set} & \textbf{Filter} &  \textbf{mAP \%}  \\\hline
			web images + videos & filtered videos & 58.11 \\ 
			web images + videos & Random Walk & \textbf{60.89} \\ \hline
		\end{tabular}
		\caption{Untrimmed video classification results on Thumos'14 dataset.}
		\label{table:ThumosRec}
	}
\end{table}

To show that the filter bias is not specific to trimmed video flow classification only, we show an additional experiment, evaluating on the untrimmed Thumos'14 dataset using an RGB network. Here we train on web images and video frames jointly but we compare two methods of selecting frames from the video. One is by using the RGB image network to identify confident frames, the other performs a random walk on images and video frames jointly. Again, we observe degraded performance due to filter bias.

\subsection{Evaluation of Fusion Schemes}
In addition, we analyze the performance of the introduced fusion schemes, \ie fusion-by-average and fusion-by-product, and the influence of the number of stacked flow maps $D$ input into the flow network in Table \ref{table:RGBnFLOW}. 

\begin{table}
	\centering
	\begin{tabular}{  lll | r }\hline
		\textbf{Input} & & \textbf{Fusion} &\textbf{Accuracy \%} \\ \hline
		Flow & D=1 & - & 50.61 \\ 
		Flow & D=10& - & 51.52 \\
		Flow & (D=1) + (D=10) & average & 52.62 \\ \hline
		RGB + Flow & D=1 & average & 71.24 \\ 
		RGB + Flow & D=1 & product & 72.75 \\ \hline
		RGB + Flow & (D=1) + (D=10) & product & \textbf{74.7} \\\hline
	\end{tabular}
	\caption{Analyzing the influence of the number of stacked flow maps $D$ and the two fusion methods on UCF-101.}
	\label{table:RGBnFLOW}
\end{table}

We show that in the end combining all three networks - both flow and the RGB one - yields the best results. 
Combining both flow networks is beneficial since short ($D=1$) and longer ($D=10$) temporal dependencies can be captured. 
When adding the RGB network, product fusion between different modalities seems to emphasize their synergies. 
The overall impact of adding temporal information improves the recognition accuracy for 88 out of the 101 classes of UCF-101. 


\subsection{Action Classification of Trimmed and Untrimmed Videos}\label{exp:RGBNets}
We compare our approach to state-of-the-art methods that use training data (Table \ref{table:oursVsLabeled}) as well as those being purely webly-supervised (Table \ref{table:oursVsWebly}). Relevantly, without using even a single humanly annotated training sample, our approach performs better than \cite{karpathy2014large, donahue2015long} who use annotations for training. In comparison to webly-supervised approaches, our CNN approach works slightly better than Lead-Exceed network without LSTM \cite{gan2016you}.

\begin{table}
	\parbox{\linewidth}{
		\centering
		\begin{tabular}{l|c}\hline
			\textbf{Method}&\textbf{Accuracy \%}\\
			\hline
			Karpathy \etal \cite{karpathy2014large} & 65.4 \\
			LRCN~\cite{donahue2015long} & 71.1 \\
			LSTM composite model~\cite{srivastava2015unsupervised} & 75.8 \\
			C3D~\cite{tran2015learning} & 82.3 \\
			Two-stream network~\cite{simonyan2014twos} & \textbf{88.0} \\
			\hline
			RGB+Flow (ours) & 74.7 \\\hline
		\end{tabular}
		\caption{Comparison with state-of-the-art methods that use labeled UCF-101 dataset for training.}
		\label{table:oursVsLabeled}
	}
\end{table}

\begin{table}
	\parbox{\linewidth}{
		\centering
		\begin{tabular}{l|c}\hline
			\textbf{Method}&\textbf{Accuracy \%}\\
			\hline
			Gan \etal \cite{gan2016webly} & 69.3 \\
			Lead-Exceed (w/o LSTM)~\cite{gan2016you} &74.4 \\
			\hline
			RGB+Flow (ours) & \textbf{74.7} \\ \hline
		\end{tabular}
		\caption{Comparison with webly-supervised state-of-the-art methods tested on UCF-101.}
		\label{table:oursVsWebly}
	}
\end{table}
\begin{table}
	\parbox{\linewidth}{
		\centering
		\begin{tabular}{c|c}
			\hline
			\textbf{Method}&\textbf{mAP \%}\\
			\hline
            		Jain \etal \cite{jain2014university} & \textbf{69.3} \\ 
            		INRIA Lear~\cite{oneata2014lear} & 64.4 \\ 
			\hline
			ours & 60.8 \\ 
			\hline
		\end{tabular}
		\caption{Thumos'14 results and the state of the art.}
		\label{table:oursVsallThumos}
	}
\end{table}

For the case of action recognition in un-trimmed video on Thumos'14, to the best of our knowledge, we do not have a direct comparison in webly-supervised methods doing un-trimmed recognition. Table \ref{table:oursVsallThumos} shows a comparison of our approach against the methods, both trained on the annotated dataset from the Thumos'14 challenge.

\subsection{Action Localization Results}
As previously described, the aim of the action localization task is to recognize and localize an action temporally in a given untrimmed video, i.e. the output is represented by a real-valued score indicating the confidence of the prediction, together with the starting and ending frame for the given action. We report experiments for both introduced localization schemes, i.e. \emph{Frame-by-frame} and \emph{Sliding window} where we report results for different window sizes.

\begin{table}[ht]
		\parbox{\linewidth}{
		\centering
	\begin{tabular}{lccccc}
		\hline
		\textbf{Overlap Ratio} &0.1&0.2&0.3
		&0.4&0.5\\ \hline
		\textbf{Supervised SotA} &&&&&\\
		Wang \etal \cite{wang2014action} & 18.1 & 17.0 & 14.1 & 11.7 & 8.3\\
		Oneata \etal \cite{oneata2014lear} & 36.6 & 33.6 & 27.0 & 20.8 &14.4
		\smallskip \\
		\textbf{Webly SotA} &&&&&\\
		Sun \etal \cite{sun2015} & 12.4 & 11.0 & 8.5 & 5.2 & 4.4
		\smallskip \\
		\textbf{Ours} &&&&&\\
		Frame-by-frame & 20.5 & 16.5 & 11.2 & 6.8 & 3.7\\
		Sliding window & 15.4 & 13.1 & 9.1 & 5.4 & 2.6\\ \hline
	\end{tabular}
	\caption{Results on Thumos'14 test set for temporal localization in untrimmed videos and comparison with the state-of-the-art.}
	\label{table:localization}
	}
\end{table}

It can be seen that the frame-by-frame method of prediction works better than the sliding window approach in these experiments. This may be due to over-estimating the temporal bounds with sliding window which causes the intersection over union score to deteriorate. The shorter actions are detected better in a frame-by-frame manner. The proposed method works better than the webly state-of-the-art for smaller overlap ratios. This can be due to detecting only the highlights, e.g.~actions present in web images, but not the full activity from beginning to end. In comparison to the state-of-the-art methods that use the manually annotated data for training, the webly supervised approach still lags behind in performance.
 
\subsection{Qualitative Analysis}
\begin{figure}[t]
	\centering
	\includegraphics[width=0.9\linewidth]{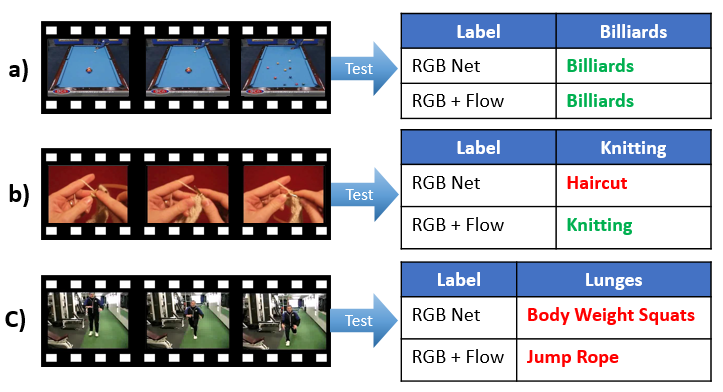}
	\caption{Some examples of test results on UCF-101 dataset}
	\label{fig:qualitative}
\end{figure}


In general, those actions that are related to a characteristic scenario or that involve a specific object tend to be accurately recognized, as they can directly be identified from individual RGB frames, which are abundantly available on the web to learn from. As an example, actions like \texttt{billiards}, \texttt{bowling} and actions associated to playing a musical instrument are all identified correctly with almost 100\% accuracy by means of RGB nets only. Recognition becomes more difficult with actions which have intricate motion, for which optical flow helps (e.g., the knitting action Figure~\ref{fig:qualitative}b). However some actions, especially those which involve body movements only, are very difficult for the model to recognize. For instance, \texttt{Jumping Jacks}, \texttt{Jump Rope} and \texttt{Lunges} (0\% recall rate) have a limited number of real world images to be trained on, and, as said before, images from Google suffer from source bias. In these cases, optical flow also does not help much due to lack of descriptive samples in the training set, often being confused by similar classes characterized by similar body movements. For instance \texttt{Lunges} is often confused with \texttt{Body weight squats} and \texttt{Jump Rope} (all characterized by similar up-and-down movements). 

\section{Conclusion}
Webly-supervised methods open a new direction to extend video analysis tasks to a larger scale and at a lower cost compared to current systems. Our experiments on two large-scale datasets demonstrate that data collected from the web is effective in training powerful models for human activity recognition in videos. Our method is even comparable to some existing approaches that rely on manually-labeled supervised training sets. Additionally, we improve over the state of the art for webly-supervised methods. We identify two biases: source bias and filter bias, that can occur in webly trained models at different stages of the pipeline, and show how they can be effectively reduced.
The results further encourage the research into the direction of webly-supervised methods, as it holds the potential to save both time and money associated with data annotation. As analyzed in details at the end of our experimental section, future research work should be targeted at improving webly recognition for actions with limited real world training data and characterized by similar body movements.
\section*{Acknowledgments}
L. Ballan was supported by the EC’s FP7 under the grant agreement No. 623930 (Marie Curie IOF).
C. Rupprecht was supported by the TUM - Institute for Advanced Study (German Excellence Initiative - FP7 Grant 291763). This work was partially funded by the Bavaria California Technology Center (BaCaTec grant No.~12[2016-1]). We tank NVIDIA for the donation of a GPU.

\bibliography{bibliography}

\end{document}